\pdfoutput=1

\documentclass[11pt]{article}

\usepackage[]{EMNLP2023}

\usepackage{times}
\usepackage{latexsym}

\usepackage[T1]{fontenc}

\usepackage[utf8]{inputenc}

\usepackage{microtype}

\usepackage{inconsolata}

\usepackage{booktabs} 
\usepackage{graphicx}  
\usepackage{multirow}
\usepackage{subfigure}
\usepackage{xspace}
\usepackage{colortbl}
\usepackage{xcolor}
\usepackage{pifont}
\usepackage{amsmath}
\usepackage{amsfonts}
\usepackage{resizegather}
\usepackage{algorithm}
\usepackage{algorithmic} 
\usepackage{amssymb}
\usepackage{svg}
\usepackage{arydshln}

\usepackage{bm}

\usepackage{tikz}
\usepackage{tikz}
\usepackage[edges]{forest}
\definecolor{hidden-draw}{RGB}{205, 44, 36}
\definecolor{hidden-blue}{RGB}{194,232,247}
\definecolor{hidden-orange}{RGB}{243,202,120}
\definecolor{hidden-yellow}{RGB}{242,244,193}
\definecolor{tree-level-1}{RGB}{245,20,85}
\definecolor{tree-level-2}{RGB}{246,86,118}
\definecolor{tree-level-3}{RGB}{248,177,193}
\definecolor{tree-leaf}{RGB}{176,230,198}

\usepackage{lscape}
\usepackage{caption}

\usetikzlibrary{positioning,shapes,shadows,arrows}
\usepackage{bbm}
\usepackage{array}
\usepackage{multirow}
\usepackage{longtable}
\usepackage{float, caption}
\usepackage{soul}
\newcolumntype{L}[1]{>{\raggedright\let\newline\\\arraybackslash\hspace{0pt}}m{#1}}
\newcolumntype{C}[1]{>{\centering\let\newline\\\arraybackslash\hspace{0pt}}m{#1}}
\newcolumntype{R}[1]{>{\raggedleft\let\newline\\\arraybackslash\hspace{0pt}}m{#1}}

\newcolumntype{x}[1]{
>{\centering\hspace{0pt}}p{#1}}%

\newcommand{\ignore}[1]{}

\usepackage{color, colortbl}
\definecolor{Gray}{gray}{0.9}
\definecolor{LightCyan}{rgb}{0.88,1,1}

\usepackage{booktabs,arydshln}

\makeatletter
\def\adl@drawiv#1#2#3{%
        \hskip.5\tabcolsep
        \xleaders#3{#2.5\@tempdimb #1{1}#2.5\@tempdimb}%
                #2\z@ plus1fil minus1fil\relax
        \hskip.5\tabcolsep}
\newcommand{\cdashlinelr}[1]{%
  \noalign{\vskip\aboverulesep
           \global\let\@dashdrawstore\adl@draw
           \global\let\adl@draw\adl@drawiv}
  \cdashline{#1}
  \noalign{\global\let\adl@draw\@dashdrawstore
           \vskip\belowrulesep}}
\makeatother

\usepackage[hang]{footmisc}
\title{Leveraging Large Language Models for NLG Evaluation: \\Advances and Challenges}

\author{Zhen Li$^{\clubsuit}$\thanks{\ \ Equal Contribution.}\ \ , \textbf{Xiaohan Xu}$^{\bigtriangleup}$\footnotemark[1]\ \ , {Tao Shen}$^{\diamondsuit}$,  Can Xu$^{\clubsuit}$, Jia-Chen Gu$^{\heartsuit}$, Yuxuan Lai$^{\bigtriangledown}$, \\
\textbf{Chongyang Tao}$^{\spadesuit}$\thanks{\ \ Corresponding author.} ,  \textbf{Shuai Ma}$^{\spadesuit}$ \\
 $^{\spadesuit}$SKLSDE Lab, Beihang University  \quad 
$^{\clubsuit}$Peking University  \\
$^{\bigtriangleup}$Institute of Information Engineering, CAS \\
$^{\diamondsuit}$AAII, FEIT, University of Technology Sydney \quad 
$^{\heartsuit}$UCLA  \quad $^{\bigtriangledown}$The Open University of China \\
{\tt lizhen63@pku.edu.cn} \quad {\tt xuxiaohan@iie.ac.cn} \quad
{\tt tao.shen@uts.edu.au} \\
{\tt gujc@ucla.edu} \quad  {\tt \{chongyang,mashuai\}@buaa.edu.cn}   
 \\
}

\begin{document}
\maketitle
\begin{abstract}

In the rapidly evolving domain of Natural Language Generation (NLG) evaluation, introducing Large Language Models (LLMs) has opened new avenues for assessing generated content quality, e.g., coherence, creativity, and context relevance. This paper aims to provide a thorough overview of leveraging LLMs for NLG evaluation, a burgeoning area that lacks a systematic analysis. We propose a coherent taxonomy for organizing existing LLM-based evaluation metrics, offering a structured framework to understand and compare these methods. Our detailed exploration includes critically assessing various LLM-based methodologies, as well as comparing their strengths and limitations in evaluating NLG outputs. By discussing unresolved challenges, including bias, robustness, domain-specificity, and unified evaluation, this paper seeks to offer insights to researchers and advocate for fairer and more advanced NLG evaluation techniques.

\end{abstract}

\section{Introduction}

Natural Language Generation (NLG) stands at the forefront of modern AI-driven communication, with recent advancements in large language models (LLMs) revolutionizing the capabilities of NLG systems~\cite{ouyang2022training, openai2023gpt4}. These models, powered by deep learning techniques and vast amounts of training data, exhibit excellent proficiency in generating text across a wide range of applications. As NLG technology continues its rapid evolution, it becomes increasingly imperative to establish robust evaluation methodologies that can reliably gauge the quality of the generated content. 

Traditional NLG evaluation metrics, such as BLEU~\cite{Papineni2002Bleu}, ROUGE~\cite{Lin2004Rouge} and TER~\cite{snover2006study}, primarily focus on surface-level text differences and often fall short in assessing semantic aspects~\cite{freitag-etal-2020-bleu}. This limitation has been noted to hinder research progress and can lead to misleading research conclusions. 
Additionally, other methods that employ neural embeddings to calculate the score~\cite{liu2016not,sellam-etal-2020-bleurt,Zhang2020BERTScore}, despite assessing aspects like semantic equivalence and fluency, are inflexible and limited in scope~\cite{freitag-etal-2021-experts}. 
Additionally, these traditional methods tend to have low alignment with human judgement~\cite{liu2023gpteval} and lack interpretability for the score~\cite{xu2023instructscore}.
These drawbacks underscore the need for more nuanced and comprehensive evaluation methods in the NLG field.

\begin{figure}[t!]
  \centering   
  \includegraphics[width=0.5\textwidth]{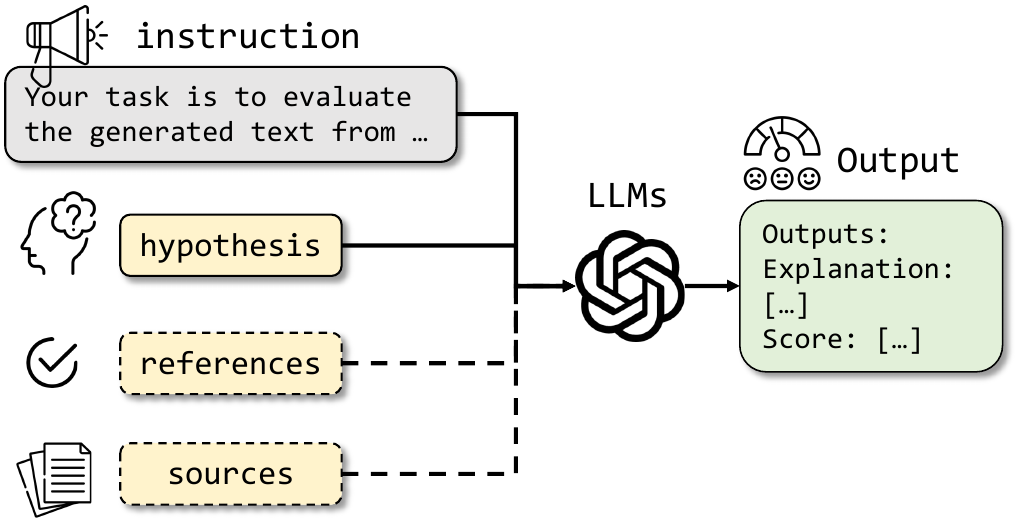}
  \caption{Illustration of LLMs for NLG evaluation. The dashed line means that the references and sources are optional based on the scenarios.}
  \vspace{-3mm}
  \label{fig:illu}
\end{figure}

The emergent abilities of LLMs present a promising avenue for the LLM-based NLG evaluation, such as Chain-of-Thought (CoT)~\cite{wei2022chain}, zero-shot instruction following~\cite{DBLP:conf/iclr/WeiBZGYLDDL22}, better alignment with human preference~\cite{ouyang2022training}, etc. These attributes position LLMs as potent tools for evaluating NLG outputs, offering a more sophisticated and better human-aligned assessment compared to traditional methods~\cite{liu2023gpteval, kocmi-federmann-2023-large, fu2023gptscore}. 
For instance, LLMs could generate reasonable explanations to support the ultimate score~\cite{xu2023instructscore}, and the reinforcement learning with human feedback (RLHF) could align LLMs' preference with human better~\cite{ouyang2022training, zheng2023judging}.
As in Figure~\ref{fig:illu}, the key strategy in these approaches involves instructing LLMs with prompts to evaluate generated texts from various aspects, either with references and sources or not. However, the wide array of LLM-based NLG evaluation methods, addressing different tasks and goals, lack a unified overview.

Given the burgeoning volume of work in the realm of LLMs for NLG evaluation, a synthesized summary is urgently needed to navigate the complexities and diverse methodologies within this space. This survey aims to provide a comprehensive overview of this promising domain, presenting a coherent taxonomy for organizing existing works. We meticulously delineate pivotal studies and their methodologies, and delve into an analytical discussion of the various strengths, limitations, and distinctive attributes of these approaches. Furthermore, we navigate through the yet-to-be-resolved challenges and the open-ended questions within this field, thereby charting potential avenues for future scholarly exploration. This comprehensive exploration aims to spark readers with an in-depth understanding of the nuances and evolving dynamics of LLM-based approaches in NLG evaluation.

\textbf{Organization of this paper:} We present the first comprehensive survey of recent advancements in leveraging LLMs for NLG evaluation. Initially, we establish a formal framework for NLG evaluation and propose a taxonomy to categorize relevant works (Section \ref{sec:taxonomy}). Subsequently, we delve into and elaborate on these works in detail (Section \ref{sec:generative}). Furthermore, we conduct a systematic review of various meta-evaluation benchmarks that assess the efficacy of LLM-based evaluators (Section \ref{sec:benchmark}). Additionally, we provide a thorough comparison of LLM-based evaluators with traditional evaluators in terms of performance, efficiency and qualitative qualitative analysis  (Section \ref{sec:comparison}).
In recognition of the rapid evolution of this field, we identify and discuss several potential open problems that may guide future research (Section \ref{sec:future}). To conclude, we advocate for the advancement of this field through the development of more impartial, robust, expert and unified LLM-based evaluators. 

\section{Formalization and Taxonomy}
\label{sec:taxonomy}

\begin{figure*}[t!]
  \centering   
  \includegraphics[width=0.78\textwidth]{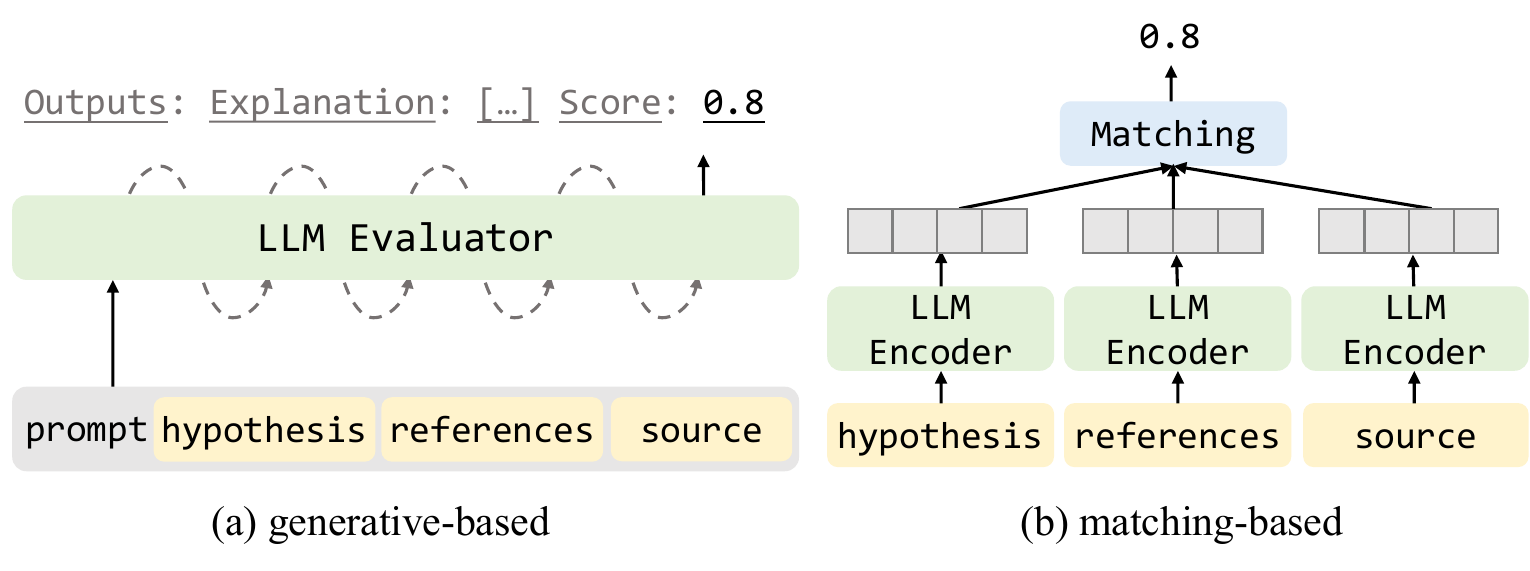}
  \caption{Illustration of NLG evaluation functions: (a) generative-based and (b) matching-based methods.}
  \vspace{-2.5mm}
  \label{fig:fine}
\end{figure*}

\tikzstyle{my-box}=[
    rectangle,
    draw=hidden-draw,
    rounded corners,
    text opacity=1,
    minimum height=1.5em,
    minimum width=5em,
    inner sep=2pt,
    align=center,
    fill opacity=.5,
]
\tikzstyle{leaf}=[my-box, minimum height=1.5em,
    fill=hidden-orange!60, text=black, align=left,font=\scriptsize,
    inner xsep=2pt,
    inner ysep=4pt,
]
\begin{figure*}[t]
    \centering
    \resizebox{\textwidth}{!}{
        \begin{forest}
            forked edges,
            for tree={
                grow=east,
                reversed=true,
                anchor=base west,
                parent anchor=east,
                child anchor=west,
                base=left,
                font=\small,
                rectangle,
                draw=hidden-draw,
                rounded corners,
                align=left,
                minimum width=4em,
                edge+={darkgray, line width=1pt},
                s sep=3pt,
                inner xsep=2pt,
                inner ysep=3pt,
                ver/.style={rotate=90, child anchor=north, parent anchor=south, anchor=center},
            },
            where level=1{text width=4.3em,font=\scriptsize,}{},
            where level=2{text width=5.8em,font=\scriptsize,}{},
            where level=3{text width=6.1em,font=\scriptsize,}{},
            where level=4{text width=6.1em,font=\scriptsize,}{},
            [ LLMs for NLG \\ Evaluation
                [
                    Taxonomy \\ of Generative \\ Evaluation (\S \ref{sec:generative})
                    [
                        Prompt-based (\S \ref{sec:prompt})\\
                        [
                            Score-based \\ 
                            [
                                    GEMBA~\cite{kocmi-federmann-2023-large}{,}
                                    Lin~\cite{lin2023llm}{,}
                                    Liu~\cite{liu2023evaluate}{,} \\
                                    Wang~\cite{wang2023chatgpt}{,}
                                    ICE~\cite{jain2023multi}{,}
                                    Embed Llama~\cite{dreano-etal-2023-embed} 
                                    , leaf, text width=26em
                            ]
                        ]
                        [
                            Probability-based \\ 
                            [
                                    BARTSCORE~\cite{Yuan2021BARTScore}{,}
                                    GPTSCORE~\cite{fu2023gptscore}{,}
                                    FFLM~\cite{jia2023zero}
                                    , leaf, text width=26em
                            ]
                        ]
                        [
                            Likert-style \\ 
                            [
                                    GEMBA~\cite{kocmi-federmann-2023-large}{,}
                                    Luo~\cite{luo2023chatgpt}{,}
                                    Gao~\cite{gao2023human}{,} \\
                                    Skopek~\cite{skopek2023towards}{,}
                                    LLM-ToT-eval~\cite{zhao2023investigating}{,}
                                    Attrscore~\cite{yue2023automatic}{,} \\
                                    Chen~\cite{chen2023exploring}{,}
                                    Bai~\cite{bai2023benchmarking}{,} 
                                    Gilardi~\cite{gilardi2023chatgpt}{,} \\
                                    Huang~\cite{huang2023chatgpt}{,}
                                    LLM-longeval~\cite{wu2023less}{,}
                                    LLM-judge~\cite{zheng2023judging}{,} \\
                                    Zhuo~\cite{zhuo2023large}{,}
                                    Sottana~\cite{sottana2023evaluation}{,}
                                    Ostheimer~\cite{ostheimer2023text}{,} \\
                                    AUTOCALIBRATE~\cite{liu2023calibrating}{,}
                                    Chiang~\cite{chiang-lee-2023-large}
                                    , leaf, text width=26em
                            ]
                        ]
                        [
                            Pairwise  \\ 
                            [
                                    Luo~\cite{luo2023chatgpt}{,}
                                    Gao~\cite{gao2023human}{,}
                                    FairEval~\cite{wang2023large}{,} 
                                    Ji~\cite{ji2023exploring}{,} \\
                                    LLM-judge~\cite{zheng2023judging}{,}
                                    EvalLM~\cite{kim2023evallm}{,} 
                                    Bai~\cite{bai2023benchmarking}{,} \\
                                    Chen~\cite{chen2023exploring}{,}
                                    AuPEL~\cite{wang2023automated} 
                                    , leaf, text width=26em
                            ]
                        ]
                        [
                            Ensemble \\ 
                            [
                                    DRPE~\cite{wu2023large}{,}
                                    WideDeep~\cite{zhang2023wider}{,}
                                    ChatEval~\cite{chan2023chateval}{,} \\
                                    Prd~\cite{li2023prd}
                                    , leaf, text width=26em
                            ]
                        ]
                        [
                            Advanced \\ 
                            [
                                    EAprompt~\cite{lu2023error}{,}
                                    Geval~\cite{liu2023gpteval}{,}
                                    FACTSCORE~\cite{min2023factscore}{,} \\
                                    ALLURE~\cite{hasanbeig2023allure}{,}
                                    Para-Ref~\cite{tang2023not}
                                    , leaf, text width=26em
                            ]
                        ]
                    ]
                    [
                        Tuning-based (\S \ref{sec:tuning})\\  
                        [
                            Probability-based \\ 
                            [
                                    PRISM~\cite{thompson-post-2020-automatic}{,}
                                    T5SCORE~\cite{qin2022t5score}
                                    , leaf, text width=26em
                            ]
                        ]
                        [
                            Likert-style \\ 
                            [
                                    TrueTeacher~\cite{gekhman2023trueteacher}{,}
                                    PERSE~\cite{wang2023learning}{,}
                                    Attrscore~\cite{yue2023automatic}{,} \\
                                    AUTO-J~\cite{li2023generative}{,}
                                    Prometheus~\cite{kim2023prometheus}{,}
                                    CritiqueLLM~\cite{ke2023critiquellm} {,} \\
                                    X-EVAL~\cite{liu2023x}
                                    , leaf, text width=26em
                            ]
                        ]
                        [
                            Pairwise  \\ 
                            [
                                    PandaLM~\cite{wang2023pandalm}{,}
                                    AUTO-J~\cite{li2023generative}{,}
                                    LLM-judge~\cite{zheng2023judging}{,} \\
                                    PERSE~\cite{wang2023learning}{,}
                                    Prometheus~\cite{kim2023prometheus}
                                    , leaf, text width=26em
                            ]
                        ]
                        [
                            Advanced \\ 
                            [
                                    INSTRUCTSCORE~\cite{xu2023instructscore}{,}
                                    TIGERScore~\cite{jiang2023tigerscore}
                                    , leaf, text width=26em
                            ]
                        ]
                    ]
                ]
                [
                    Meta-Evaluation \\ Benchmarks (\S \ref{sec:benchmark})
                    [
                        Machine Translation
                        [
                            MQM~\cite{freitag-etal-2021-experts}{,}
                            WMT Metrics Shared Task~\cite{mathur-etal-2020-results,freitag-etal-2021-results,freitag-etal-2022-results}
                            , leaf, text width=33.6em
                        ]
                    ]
                    [
                        Text Summarization
                        [
                            NEWSROOM~\cite{grusky-etal-2018-newsroom}{,}
                            SamSum~\cite{gliwa2019samsum}{,}
                            REALSumm~\cite{bhandari-etal-2020-evaluating}{,} \\
                            QAGS\_XSUM ~\cite{Wang2020QAGS}{,} 
                            FRANK~\cite{pagnoni-etal-2021-understanding}{,}
                            SummEval~\cite{fabbri-etal-2021-summeval}{,} \\
                            SummaC~\cite{laban-etal-2022-summac}{,} 
                            RiSum~\cite{skopek2023towards}{,}
                            OpinSummEval~\cite{shen2023opinsummeval}
                            , leaf, text width=33.6em
                        ]
                    ]
                    [
                        Dialogue Generation
                        [
                            FED~\cite{mehri-eskenazi-2020-unsupervised}{,}
                            Topical-Chat~\cite{Gopalakrishnan2019TopicalChatTK}{,}
                            PersonaChat~\cite{zhang2018personalizing}
                            , leaf, text width=33.6em
                        ]
                    ]
                    [
                        Image Caption
                        [
                            Flickr8K-Expert~\cite{hodosh2013framing}{,}
                            Composite~\cite{aditya2015images}{,}
                            Pascal-50S~\cite{vedantam2015cider}{,} \\
                            MSCOCO Image Captioning Challenge~\cite{cui2018learning}
                            , leaf, text width=33.6em
                        ]
                    ]
                    [
                        Data-to-Text
                        [
                            BAGEL~\cite{mairesse-etal-2010-phrase}{,}
                            SFRES~\cite{wen-etal-2015-semantically}{,}
                            SFHOT~\cite{wen-etal-2015-semantically}{,}
                            WebNLG~\cite{castro-ferreira-etal-2020-2020}
                            , leaf, text width=33.6em
                        ]
                    ]
                    [
                        Story Generation
                        [
                            OpenMEVA~\cite{guan-etal-2021-openmeva}{,}
                            WP$_{200}$~\cite{chen-etal-2022-storyer}{,}
                            SCARY$_{200}$~\cite{chen-etal-2022-storyer}{,}
                            PREF$_{200}$~\cite{chen-etal-2022-storyer}{,} \\
                            COH$_{200}$~\cite{chen-etal-2022-storyer}{,}
                            Per-MPST~\cite{wang2023learning}{,}
                            Per-DOC~\cite{wang2023learning}
                            , leaf, text width=33.6em
                        ]
                    ]
                    [
                        General Generation
                        [
                            AlpacaEval~\cite{alpaca_eval}{,}
                            MT-bench~\cite{zheng2023judging}{,}
                            FairEval~\cite{wang2023large}{,}
                            Shepherd~\cite{wang2023shepherd}{,} \\
                            LLMBar~\cite{zeng2023evaluating}{,} 
                            LLMeval~\cite{zhang2023wider}{,} 
                            AttrEval-GenSearch~\cite{yue2023automatic}{,} \\
                            ALIGNBENCH~\cite{liu2023alignbench}
                            , leaf, text width=33.6em
                        ]
                    ]
                ]
            ]
        \end{forest}
    }
    \caption{Taxonomy of research in NLG evaluation with large language models.}
    \label{overall}
\end{figure*}
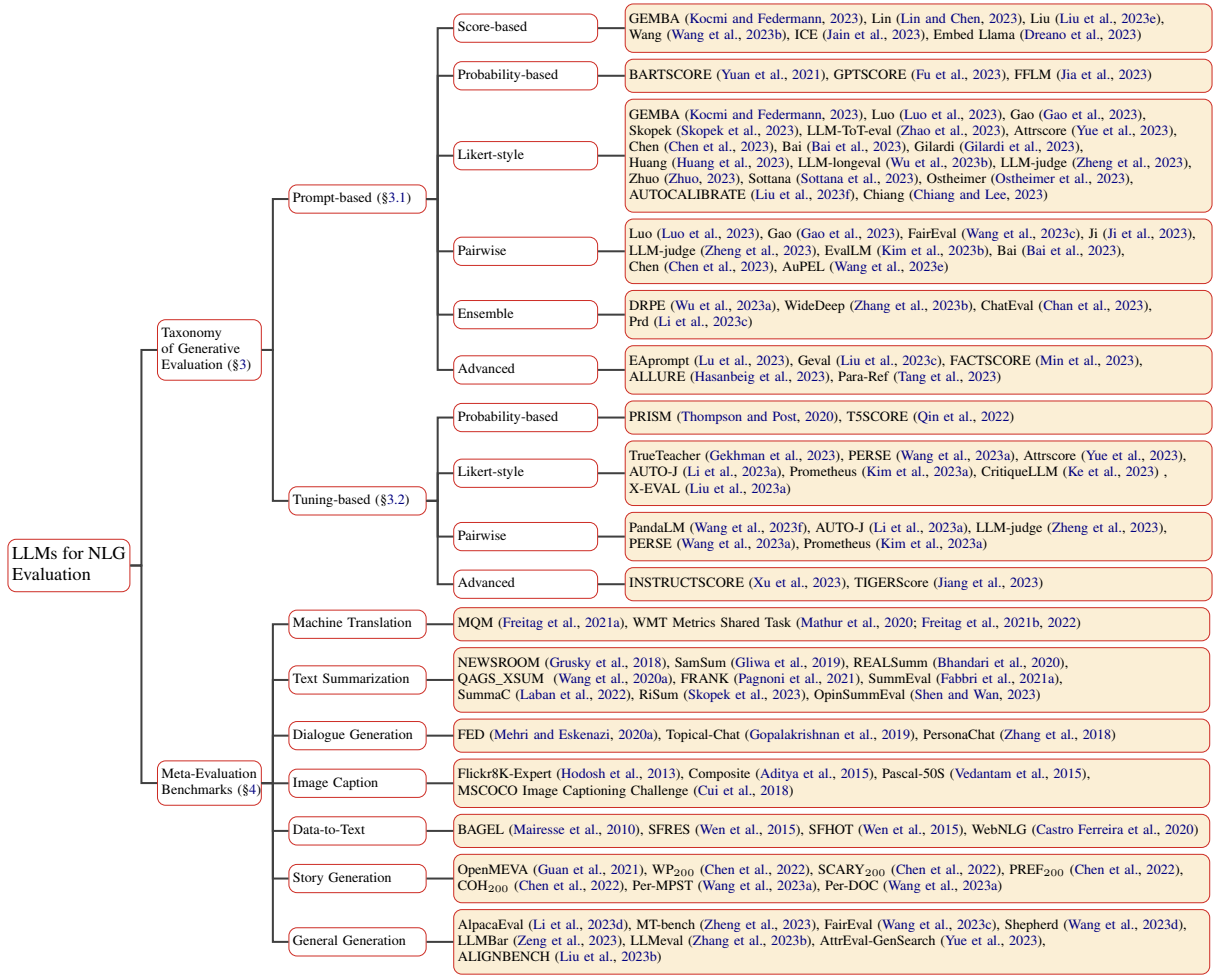

In this section, we first briefly formalize LLM-based NLG Evaluation tasks.  
The objective of NLG evaluation is to assess the candidate generations of a model across various dimensions, such as fluency, consistency, etc. 
Recent advancements in LLMs have significantly enhanced their capabilities in context comprehension and the generation of reasonable responses. 
Notably, contemporary research has begun to reframe NLG evaluation as a series of instruction-following tasks, leveraging powerful capabilities of LLMs~\cite{Zhang2020BERTScore, fu2023gptscore}. 
To maintain generality, we formalize the existing evaluation framework for texts generated by models as follows:
\begin{equation} \label{eq:formalization}
    E = f({h}, {s}, {r}). 
\end{equation}
Here, $h$ represents the hypothesis text (i.e. candidate generation to be evaluated), and $f$ denotes the evaluation function, which can be instantiated by LLMs. The variable $s$ denotes the input source of the generation. This source might include the source text or any supporting documents that provide background or framing for the generated content. For instance, in machine translation tasks, $c$ could be the sentence in the source language. Lastly, $r$ refers to a set of ground truth references that serve as a benchmark for evaluation. These references are crucial in tasks like text summarization, where the quality of the generated summary is assessed by an annotated reference summary.

In this paper, we classify works in NLG evaluation along three primary dimensions: \textit{evaluation task}, \textit{evaluation references} and \textit{evaluation function}. These dimensions provide a comprehensive framework for categorizing and understanding different approaches within this domain.

\paragraph{Evaluation Task $\mathcal{T}$}: 
NLG encompasses a diverse range of tasks, such as Machine Translation (MT)~\cite{farhad2021findings,bapna2019simple}, Text Summarization (TS)~\cite{liu-liu-2021-simcls,zhang2023summit}, Dialogue Generation (DG)~\cite{tao2018ruber,kann2022open}, Story Generation (SG)~\cite{yang2022re3,fan-etal-2018-hierarchical}, Image Caption (IC)~\cite{tewel2022zerocap,zhou2022towards}, Data-to-Text generation (D2T)~\cite{lin2023survey,jing2023stylized} and General Generation (GE)~\cite{wang-etal-2023-self-instruct,zheng2023judging}, each with its unique evaluation requirements and challenges. The specific nature of each task determines the target evaluation aspects and scenarios.
For instance, in text summarization, the focus may be on relevance to the source content, while in dialogue generation, the coherence of the response is crucial.
Given these varied objectives, our taxonomy also extends to the lens of task-specific evaluation.
This categorization enables us to understand how different evaluation methods perform across a spectrum of NLG tasks, thus offering insights into the strengths and limitations of existing evaluation paradigms in distinct task contexts.

\paragraph{Evaluation References $r$}: According to whether the references are available, we divide the evaluation scenarios into \emph{reference-based} and \emph{reference-free} scenarios.
In \emph{reference-based} evaluation, the generated text $h$ is compared against a set of ground truth references $r$. 
This approach is particularly prevalent in tasks where the quality of the generated text can be objectively measured against established standards. 
The comparison metrics might focus on aspects like accuracy, relevance, coherence, and the degree of similarity to the references. 
Typical applications include text summarization, where the generated summaries are evaluated against reference summaries, and machine translation, where the translations are compared with standard translations.
The \emph{reference-free} approach, in contrast, does not rely on any external references for evaluation. This method evaluates the generated text $h$ based on the intrinsic qualities or its alignment with the provided source context $s$. 
Evaluation in this context may focus on aspects such as fluency, originality, relevance to the context, etc.

\paragraph{Evaluation Function $f$}: Evaluation function could be matching-based or generative-based on the basis of different ways of utilizing LLMs. As shown in Figure~\ref{fig:fine}, {\emph{matching-based methods} measure the semantic equivalence between the reference and hypothesis or measure the proper degree between the source text and hypothesis. 
Several works measure the semantic equivalence between the reference and hypothesis by using token-level matching functions in distributed representation space~\cite{Zhang2020BERTScore, Zhao2019MoverScore} or discrete string space~\cite{Lin2004Rouge, Papineni2002Bleu}. Others focus on sequence-level, such as~\cite{sellam-etal-2020-bleurt, rei-etal-2020-comet}. }
In contrast, \emph{generative-based methods} include methods where LLMs are employed to generate evaluation metrics directly. These methods leverage the generative capabilities of LLMs to assess the quality of generated text by designing instructions.

\paragraph{\textbf{Scope of this paper:}} { Recent matching-based methods are based on a neural encoder to calculate a score-specific aspect of evaluation. However, these methods often face challenges such as limited interpretability, lower correlation with human judgments, and a restricted range of evaluated aspects~\cite{xu2023instructscore, fu2023gptscore}.}
Fortunately, the emerging capabilities of LLMs open up a wealth of possibilities for NLG evaluation. This includes improved interpretability through CoT, higher customization via instruction-following capabilities, and better alignment with human evaluations through RLHF~\cite{xu2023instructscore, zheng2023judging}.
\emph{ Given the abundance of recent surveys primarily focusing on matching-based evaluation methods (refer to~\cite{celikyilmaz2020evaluation,sai2022survey,goyal2023systematic} for comprehensive summaries), our paper is dedicated to exploring more burgeoning generative-based methods.}
 Figure \ref{overall} presents our taxonomy of generative-based evaluation. We classify relevant works into two main categories: prompt-based and tuning-based evaluation, depending on whether the LLM is tuned. Further, we divide these methods into subcategories: score-based, probability-based, likert-style, pairwise comparison, ensemble, and advanced evaluation protocols, each distinguished by their evaluation form. These categories will be detailed in Section \ref{sec:generative}.

\section{Generative Evaluation}
\label{sec:generative}

\begin{table*}[t!]
\small
\resizebox{\textwidth}{!}{
\begin{tabular}{|l|l|l|}
\hline
Prompt Type  & \multicolumn{1}{c|}{Prompt}                                                                                                                                                                                                                            & Output                                                                                       \\ \hline
Score-based  & \texttt{\begin{tabular}[c]{@{}l@{}}Given the source document: {[}…{]} \\ Given the model-generated text: {[}…{]}\\ Please score the quality of the generated text from 1 (worst) to 5 (best)\end{tabular}}                                          & \texttt{Scores: 2}                                                                           \\ \hline
Likert-style & \texttt{\begin{tabular}[c]{@{}l@{}}Given the source document: {[}…{]}\\ Given the model-generated text: {[}…{]}\\ Is the generated text consistent with the source document? (Answer Yes or No)\end{tabular}}                                       & \texttt{Yes}                                                                                 \\ \hline
Pairwise     & \texttt{\begin{tabular}[c]{@{}l@{}}Given the source document: {[}…{]}\\ Given the model-generated text 1: {[}…{]}\\ And given the model-generated text 2: {[}…{]}\\ Please answer which text is better-generated and more consistent.\end{tabular}} & \texttt{Text 1}                                                                              \\ \hline
\end{tabular}
}
\caption{Illustration of different types of prompts.}
\label{tab:typical_prompt}
\end{table*}

Amidst the rapid evolution of LLMs, a burgeoning body of research has directed its focus toward leveraging these models as evaluators for NLG tasks. This attention is particularly rooted in the high-capacity generative abilities of LLMs, leading to the emergence of works employing them to produce quality evaluations of NLG text—a paradigm we refer to as generative evaluation. This category, broadly classified into \emph{\textbf{prompt-based evaluation}} and \emph{\textbf{tuning-based evaluation}}, hinges on whether the parameters of LLM evaluators require fine-tuning.
Prompt-based evaluation typically involves prompting robust base LLMs to assess generated text through meticulous prompt engineering. On the other hand, tuning-based evaluation relies on open-source LLMs that are specifically calibrated for NLG evaluation. Both approaches cater to diverse evaluation protocols for measuring the quality of the generated text.

Current methods consider different scoring protocols to judge the quality of generated hypothesis text.
Some endeavors deploy LLM evaluators to yield continuous scalar scores that represent the quality of individual generated texts—termed as {\color{gray}{\Large \ding{202}}} \emph{score-based evaluation}. Others calculate the generation probability of generated texts based on prompts, sources, or reference texts (optional) as the evaluation metric, denoted as {\color{gray}{\Large \ding{203}}} \emph{probability-based evaluation}. Further diversifying the landscape, certain works transform NLG evaluation into a classification task by categorizing text quality into multiple levels using likert scales. In this scenario, LLM evaluators assess the quality of generated text by assigning it to a specific quality level—referred to as {\color{gray}{\Large \ding{204}}} \emph{likert-style evaluation}. Meanwhile, {\color{gray}{\Large \ding{205}}} \emph{pairwise comparison methods} involve using LLM evaluators to compare the quality of pairs of generated texts. Additionally, {\color{gray}{\Large \ding{206}}} \emph{ensemble evaluation methods} utilize multiple LLM evaluators with different LLMs or prompts, orchestrating communication among evaluators to yield final evaluation results. 
Finally, some recent studies explore {\color{gray}{\Large \ding{207}}} \emph{advanced evaluation methods} (that consider fine-grained criteria or combine the capabilities of chain-of-thought or in-context leaning) with the goal of attaining more comprehensive and nuanced evaluation results.

This section delves into a detailed exploration of these two overarching categories of evaluation methods, each accompanied by their respective evaluation protocols. Table~\ref{tab:auto_metric_adoption} provides a comprehensive overview of current prompt-based and tuning-based evaluation methods. The table delineates their respective adaptation tasks, backbone models, scoring protocols, and evaluated aspects for clarity and reference.

\subsection{Prompt-based Evaluation}
\label{sec:prompt}

Prompt-based text evaluation stands at the forefront of advancements in NLG, particularly leveraging the capabilities of LLMs. 
In this method, the evaluation process is intricately woven into the crafting of prompts -- specialized cues designed to guide LLMs in assessing the quality and coherence of generated text.
More recently, the Eval4NLP workshop held a shared task on prompting LLMs as explainable metrics~\cite{leiter2023eval4nlp}.
Typically, a prompt template serves as a structured framework that encompasses \emph{instructions, aspects, criteria, and desired output formats}, providing a systematic guide for evaluating generated text. These templates empower researchers and practitioners to articulate precise evaluation requirements, ensuring consistency and reproducibility in the assessment process.
By harnessing the prowess of LLMs, prompt-based evaluation not only provides a comprehensive understanding of NLG system performance but also offers a nuanced approach to extracting valuable insights.

\begin{table*}[t!]
    \centering
    \resizebox{\textwidth}{!}{
    \begin{tabular}{cp{0.25cm}p{0.25cm}p{0.25cm}p{0.25cm}p{0.25cm}p{0.25cm}p{0.25cm}p{0.3cm}cp{1.4cm}c}
        \toprule
         Metric & MT & TS & DG & IC & D2T & SG & GE & REF & LLMs	& Protocol	& Aspects\\
         \midrule
         \multicolumn{12}{c}{ \emph{Prompt-based Evaluation}}\\
         \midrule

         BARTScore~\cite{Yuan2021BARTScore} & \checkmark  & \checkmark & * & * & \checkmark & * & * & \checkmark & BART & Prob & 
         \begin{tabular}[c]{@{}l@{}} CON/COH/REL/FLU/ \\ \qquad INF/COV/ADE \end{tabular} 
         \\ 
         \cdashlinelr{1-12}

         GPTScore~\cite{fu2023gptscore} & \checkmark & \checkmark  & \checkmark &  & \checkmark & * & * &  & GPT3 & Prob & 
         \begin{tabular}[c]{@{}l@{}} CON/COH/REL/FLU/COV/ACC \\
         MQM/INF/FAC/INT/ENG/NAT \end{tabular} 
         \\  
         \cdashlinelr{1-12}
         
         G-EVAL~\cite{liu2023gpteval} & * & \checkmark & \checkmark & & * & * & * & & ChatGPT/GPT-4 & Advanced & 
         \begin{tabular}[c]{@{}l@{}}  CON/COH/REL/FLU \\ \quad /NAT/ENG/GRO \end{tabular} 
         \\ 
         \cdashlinelr{1-12}
         Embed Llama~\cite{dreano-etal-2023-embed} & \checkmark & * & * &  & * & * & * & & LlaMA-2 & Score & NONE \\

         ICE~\cite{jain2023multi} & * & \checkmark & * & & * & * & * & & GPT-3 & Score & CON/COH/REL/FLU\\

         GEMBA~\cite{kocmi-federmann-2023-large} & \checkmark & * & * &  & * & * & * &  & ChatGPT & Score/Likert & NONE\\
         LLM\_eval~\cite{chiang-lee-2023-large} & * & * & * &  & * & \checkmark & * & & ChatGPT & Likert & GRAM/COH/REL/LIK\\
         FairEval~\cite{wang2023large} & * & * & * & & * & * & \checkmark & & ChatGPT/GPT-4 & Pairwise & NONE\\
         AuPEL~\cite{wang2023automated} & * & * & * & & * & * & \checkmark & & PaLM-2 & Pairwise & PER/QUA/REL \\
         DRPE~\cite{wu2023large} & * & \checkmark & * & * & * & * & * & \checkmark & GPT-3 & Ensemble & CON/COH/REL/FLU/INT/USE \\

         ChatEval~\cite{chan2023chateval} & * & * & \checkmark & & * & * & \checkmark & & ChatGPT/GPT-4 & Ensemble & NAT/COH/ENG/GRO\\
         WideDeep~\cite{zhang2023wider} & * & * & * &  & * & * & \checkmark & & ChatGPT & Ensemble & COH/REL/HARM/ACC\\
         \cdashlinelr{1-12}
         
         PRD~\cite{li2023prd} & * & * & * &  & * & * & \checkmark & &
         \begin{tabular}[c]{@{}l@{}} \ \ GPT-4/GPT-3.5  \\ Vicuna/Claude/Bard \end{tabular} 
         & Ensemble & INF/COH \\
         \cdashlinelr{1-12}
         
         FACTSCORE~\cite{min2023factscore} & & * & & & & & \checkmark & & ChatGPT & Advanced & FAC\\
         
         EAprompt~\cite{lu2023error} & \checkmark & * & * &  & * & * & * & & ChatGPT/text-davinci-003 & Advanced & NONE \\
         AUTOCALIBRATE~\cite{liu2023calibrating} & * & \checkmark & * & & * & * & * & & GPT-4 & Likert & CON/COH/REL/FLU/INF/NAT \\
         ALLURE~\cite{hasanbeig2023allure} & * & \checkmark & * &  & * & * & \checkmark & & GPT-4 & Advanced & CON/COH/FLU/REL \\
         
         \midrule
         \multicolumn{12}{c}{ \emph{Tuning-based Evaluation} }\\
         \midrule

        PRISM~\cite{thompson-post-2020-automatic} & \checkmark & * & * & * & * & * & * & \checkmark & Transformer & Prob & NONE \\

        T5Score~\cite{qin2022t5score} & \checkmark & \checkmark & * & * & * & *& * & \checkmark & T5 & Prob & NONE \\
        TrueTeacher~\cite{gekhman2023trueteacher} & * & \checkmark & * & & * & * & * & & T5 & Likert & CON\\
        \cdashlinelr{1-12}
        
         Attscore~\cite{yue2023automatic} & * & * & * & & * & * & \checkmark & & 
          \begin{tabular}[c]{@{}l@{}}  Roberta/T5/GPT2 \\ LLaMA/Vicuna \end{tabular} 
         & Likert & CON \\
         \cdashlinelr{1-12}

         X-EVAL~\cite{liu2023x} & * & \checkmark & \checkmark & & \checkmark & * & * & & FLAN-T5-large & Likert & \begin{tabular}[c]{@{}l@{}} DEP/LIK/UND/FLE/INF/INQ \\  INT/SPE/COR/SEM/COH/ENG \\ NAT/GRO/CON/REL/FLU \end{tabular}\\
        
         \cdashlinelr{1-12}
         AUTO-J~\cite{li2023generative} & * & * & * & & * & * & * & & LLaMA & Likert/Pairwise & 
         \begin{tabular}[c]{@{}l@{}} ACC/CLA/FEA/CRE/THO \\  \quad STR/LAY/COM/INF
         \end{tabular} 
         \\
         \cdashlinelr{1-12}

         PERSE~\cite{wang2023learning} & * & * & * & * & * & \checkmark & * & \checkmark & LLaMA & Likert/Pairwise & INT/ADA/SUR/CHA/END  \\
         
         PandaLM~\cite{wang2023pandalm} & * & * & * & & * & * & \checkmark & & LLaMA & Pairwise & CLA/COM/FOR/ADH\\
         TIGERScore~\cite{jiang2023tigerscore} & \checkmark & \checkmark & * & & \checkmark & \checkmark & \checkmark & & LLaMA & Advanced &  COH/INF/ACC/COM \\
         INSTRUCTSCORE~\cite{xu2023instructscore} & \checkmark & * & * & * & * & * & * & \checkmark & LLaMA & Advanced & NONE \\
         Prometheus~\cite{kim2023prometheus} & * & * & * &  & * & * & \checkmark & & LLaMA-2 & Likert/Pairwise & NONE \\
         CritiqueLLM~\cite{ke2023critiquellm} & * & * & * &  & * & * & \checkmark & & ChatGLM & Likert & NONE \\
         \bottomrule
    \end{tabular}
    }
    \caption{Automatic metrics proposed (\checkmark) and adopted (*) for various NLG tasks. \textbf{REF} indicate the method is source context-free. \textbf{MT}: Machine Translation, \textbf{TS}: Text Summarization, \textbf{DG}: Dialogue Generation, \textbf{IC}: Image Captioning,  \textbf{D2T}: Data-to-Text, \textbf{SG}: Story Generation,  \textbf{GE}: General Generation. 
    We adopted the evaluation aspects for different tasks from~\citet{fu2023gptscore}. Specifically, for each evaluation aspect, \textit{CON}: consistency, \textit{COH}: coherence, \textit{REL}: relevance, \textit{FLU}: fluency, \textit{INF}: informativeness, \textit{COV}: semantic coverage, \textit{ADE}: adequacy, \textit{NAT}: naturalness, \textit{ENG}: engagement, \textit{GRO}: groundness, \textit{GRAM}: grammaticality, \textit{LIK}: likability, \textit{PER}: personalization, \textit{QUA}: quality, \textit{INT}: interest, \textit{USE}: usefulness, \textit{HARM}: harmlessness, \textit{ACC}: accuracy, \textit{FAC}: factuality, \textit{ADA}: adaptability, \textit{SUR}: surprise, \textit{CHA}: character, \textit{END}: ending, \textit{FEA}: feasibility, \textit{CRE}: creativity, \textit{THO}: thoroughness, \textit{STR}: structure, \textit{LAY}: layout, \textit{CLA}: clarity, \textit{COM}: comprehensiveness, \textit{FPR}: formality, \textit{ADH}: adherence, \textit{DEP}: topic depth, \textit{UND}: understandability, \textit{FLE}: flexibility, \textit{INQ}: inquisitiveness, \textit{SPE}: specificity, \textit{COR}: correctness, \textit{SEM}: semantic appropriateness. \textit{NONE} means that the method does not specify any aspects and gives an overall evaluation. The detailed explanation of most evaluation aspect can be found in \citet{fu2023gptscore}.
    }
    \label{tab:auto_metric_adoption}
\end{table*}


\paragraph{Score Evaluation.}
An intuitive and widely employed protocol for utilizing LLM evaluators in text evaluation involves prompting these evaluators to generate a continuous score that reflects the quality of the generated text. A concrete example of such a prompt is illustrated in the first row of Table~\ref{tab:typical_prompt}. Pioneering this method, GEMBA~\cite{kocmi-federmann-2023-large} proposed the use of LLM evaluators to assign quality scores, ranging from 0 to 100, to generate translations both with and without a reference. GEMBA demonstrates the efficacy of employing GPT-3.5 or larger LLMs for translation quality evaluation, showcasing their capabilities with simple zero-shot prompts. Building on this foundation, \citet{lin2023llm} have extended score evaluation methods to broader NLG evaluation domains, aiming to enhance the alignment between LLM evaluators and manual annotators.

\citet{liu2023evaluate} tailored LLM evaluators to assess the quality of closed-end response generation, characterized by unique and correct semantic references. Their innovative approach involves prompting LLMs evaluators to generate explanatory judgments for the generated responses, subsequently extracting numerical quality scores. Similarly, \citet{wang2023chatgpt} proposed a unified prompt applicable across various NLG evaluation tasks, which directly generates quality scores for the produced texts across different evaluation aspects, both with and without reference. Additionally, \citet{jain2023multi} employed LLM evaluators with in-context examples to evaluate summarization tasks, generating numeric strings that effectively capture the quality of summarization outputs. These diverse applications underscore the versatility and adaptability of score-based evaluation methods when harnessing LLM evaluators for comprehensive NLG assessments.

\paragraph{Probability-based evaluation.}
Recognizing that the quality of the generated text is often correlated with the ease of generation by LLMs based on source or reference text, some studies adopt a unique perspective by framing the evaluation task as a conditional generation task. In this context, the generative likelihood of the produced text is calculated, serving as the score indicative of text quality, as illustrated in the second row of Table~\ref{tab:typical_prompt}.
\citet{Yuan2021BARTScore} first leveraged BART~\cite{lewis2019bart} as an evaluator to compute the probability of the generated text based on source or reference text across diverse evaluation aspects in machine translation, text summarization, and data-to-text tasks.
Expanding on this methodology, \citet{fu2023gptscore} devised prompts that encompass task descriptions and definitions of evaluation aspects, utilized to instruct an LLM-based evaluator to calculate the generation probability of generated text.
In contrast to the conventional use of generation probability as a quality score, \citet{jia2023zero} calculated the three probability changes as the reference-free metric to evaluate the faithfulness of the generated summary. These changes include the transition from generating a summary with the source document to directly generating a summary, altering the position of the source and summary, and the shift from generating a summary with the source document to generating a summary with a specific piece of a prefix.

\begin{figure*}
\centering
\begin{minipage}{.48\textwidth}
  \centering
  \includegraphics[width=.9\textwidth]{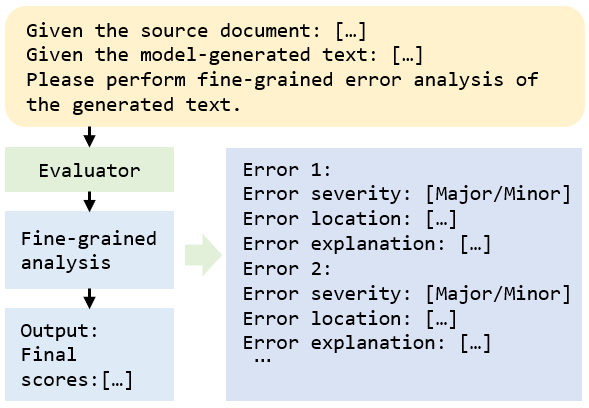}
  \captionof{figure}{A example of fine-grained evaluation inspired by \citet{jiang2023tigerscore}.}
  \label{fig:finegrain}
\end{minipage}
\hspace{2mm}
\begin{minipage}{.48\textwidth}
  \centering
  \includegraphics[width=.86\textwidth]{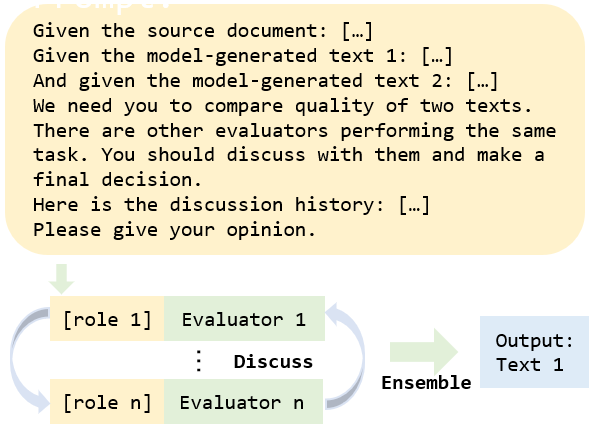}
  \captionof{figure}{A example of ensemble evaluation inspired by \citet{li2023prd}.}
  \label{fig:ensemble}
\end{minipage}
\end{figure*}

\paragraph{Likert-Style Evaluation.}
Inspired by the human annotation process, several studies employ LLM evaluators to assess the quality levels of generated texts, where these evaluators produce ratings or quality labels based on a likert-style scale~\cite{bai2023benchmarking,gao2023human,gilardi2023chatgpt,huang2023chatgpt,zhao2023investigating,wu2023less,luo2023chatgpt,zheng2023judging,zhuo2023large,sottana2023evaluation,skopek2023towards}. A representative likert-style prompt is depicted in the third line of Table~\ref{tab:typical_prompt}.
For instance, \citet{chiang-lee-2023-large} provided LLM evaluators with the same evaluation instructions as human annotators, prompting them to rate the quality of generated texts using a 5-point likert scale. Meanwhile, \citet{gao2023human} instructed ChatGPT to rate model-generated summarizations across multiple evaluation aspects, such as relevance, faithfulness, fluency, and coherence, using a scale ranging from 1 (worst) to 5 (best) based on the provided source document. In a similar vein, \citet{ostheimer2023text} designed multiple prompts, each guiding the LLM evaluator to assess a specific evaluation aspect of a text style transfer task. By comparing the transferred text with the source text, the LLM evaluator generates a discrete scale ranging from 1 to 5 to represent the quality of the transferred text. This approach exemplifies the adaptability of likert-style prompts in capturing diverse dimensions of text quality through LLM evaluations.

\citet{yue2023automatic} proposed to utilize LLM evaluator to evaluate the attribution capabilities of the generative models which judges if the generated statement is thoroughly supported by the referenced source. This work designs three categories of quality labels, including attributable, extrapolatory, and contradictory, and prompts the LLM evaluator with explicit instructions that include definitions of labels.
\citet{liu2023calibrating} utilized LLMs to draft, filter, and refine comprehensive evaluation criteria as score instructions, which achieves more consistent evaluation results with human annotators when evaluating text summarization, data-to-text generation and hallucination tasks.

\paragraph{Pairwise Evaluation.}
Compared with utilizing LLM evaluators to individually evaluate the quality of generated texts through numerical scores or likert-style rating, another way of using LLM for evaluation is to explicitly compare with other generated text and decide which one is superior~\cite{bai2023benchmarking,ji2023exploring}. A representative prompt is shown in the last row of Table~\ref{tab:typical_prompt}. 
\citet{wang2023large} utilized LLM evaluator to obtain evaluation result of two model-generated responses for one given query. This method proposes multiple evidence and balanced position calibration, and seeks assistance from human annotators when the quality of two texts is close to avoid the impact of the order of response pairs in the prompt on evaluation results.
\citet{wang2023automated} introduced a reference-free personalized text generation evaluation framework that prompts LLM evaluator to perform pairwise comparisons between the generated text pairs in three essential aspects: personalization, quality, and relevance of the generated text through providing a detailed explanation of its judgment.

\paragraph{Ensemble Evaluation.}
As the actual evaluation process often involves collaborative evaluation by multiple human annotators, some works utilize multiple LLM evaluators with different base models or prompts and allow them to evaluate the quality of generated text from different perspectives, as shown in Figure~\ref{fig:ensemble}. 
\citet{wu2023large} set multiple roles for the LLM to evaluate the quality of the generated summary by comparing it with the reference one on both subjective and objective dimensions. This work generates dynamic role profiles according to input texts and synthesizes the results of multiple roles as the final evaluation result.
\citet{li2023prd} utilized multiple LLM evaluators to conduct a pairwise evaluation for the model-generated responses by performing multiple rounds of discussions on the comparison results to reach a mutual agreement on the pairwise scoring.
Similarly, \citet{zhang2023wider} proposed to set up LLM evaluators as multiple-layer neural network structures. The bottom evaluators obtain the evaluation result of model-generated responses from a specific evaluation perspective. The upper evaluators receive all evaluation information from the previous layer and discuss it with each other to obtain a more comprehensive evaluation result.
Besides, \citet{chan2023chateval} designed diverse communication strategies with various role prompts during collaborative discussions to evaluate pairwise generated responses.

\paragraph{Advanced Evaluation.}
Some recent works investigate advanced evaluation techniques aimed at achieving more thorough and nuanced assessment outcomes by leveraging chain-of-thought, in-context learning capabilities, fine-grained analysis, etc. A representative fine-grained evaluation method is shown in Figure ~\ref{fig:finegrain}.
\citet{liu2023gpteval} utilized LLMs with chain-of-thought (CoT) prompting and a form-filling paradigm to evaluate the quality of generated texts across various NLG tasks.
\citet{min2023factscore} proposed a \textbf{\emph{find-grained evaluation schema}} that first extracts a series of atomic facts from the LLM-generated long text, and then utilizes LLM evaluator to validate each atomic fact with the given knowledge source.
\citet{lu2023error} proposed a new prompting method called Error Analysis Prompting (EAPrompt) that combines CoT to prompt the LLM evaluator to analyze different types of pre-defined errors (e.g., major and minor errors) in the generated translation based on the given source text and reference translation, and then measures the quality of a generated translation with the previous error analysis.
To enhance and improve the robustness of LLM-based evaluators,
\citet{hasanbeig2023allure} proposed ALLURE, a systematic protocol for auditing and improving LLM-based evaluation  with iterative in-context-learning.
Considering that the evaluation with a single or few references may not accurately reflect the quality of the model’s hypotheses, \citet{tang2023not} leveraged LLMs to paraphrase a single reference into multiple high-quality ones in diverse expressions, which enhances various evaluation methods on MT, TS, and caption tasks. 
To further task advantages of the in-context learning capability of LLMs, \citet{liu2023calibrating} proposed AUTOCALIBRATE to automatically align and calibrate an LLM-based evaluator through incorporating the mined and calibrated rubrics into scoring instructions.

\subsection{Tuning-based Evaluation}
\label{sec:tuning}

Recently, researchers increasingly turn their attention towards fine-tuning open-source language models (e.g., LLaMA), in lieu of traditional closed-based LLMs (e.g., ChatGPT and GPT-4). This shift is propelled by a thorough exploration of key perspectives, including the expenses associated with API calls, the robustness of prompting, and the pivotal consideration of domain adaptability.

In contrast to closed-based models that invariably demand expensive API calls for each evaluation instance, the fine-tuning of smaller open-source LLMs provides a cost-effective alternative. This approach empowers researchers to evaluate their models on specific tasks without incurring the financial burden associated with extensive API usage. Additionally, the process of prompting LLMs for NLG evaluation requires meticulous crafting of prompts, with variations potentially resulting in significant differences in outcomes.
Furthermore, the consideration of domain adaptability underscores the evolving landscape of NLG evaluation. Fine-tuning open-source LLMs affords researchers the flexibility to tailor models to diverse domains, transcending the limitations imposed by closed-based models confined to specific niches.

Typically, tuning-based methods construct evaluation data manually~\cite{zheng2023judging} or with the assistance of advanced LLMs (e.g., GPT-4)~\cite{xu2024survey}, followed by performing supervised tuning. Similar to prompting-based evaluation, tuning-based methods can also be categorized into various types based on their scoring protocol, such as \emph{likert-style evaluation}, \emph{probability-based evaluation} and \emph{pairwise evaluation}. In addition, based on the output explanations in supervised fine-tuning, these methods can be further divided into \emph{holistic evaluation} or \emph{error-oriented evaluation}. We will begin by introducing various types of scoring protocols and subsequently provide a summary of two output explanations in the final advance evaluation.

\paragraph{Likert-Style Evaluation.}
Some works tune LLMs to provide quality ratings or labels for generated texts. \citet{gekhman2023trueteacher} employed FLAN-PaLM 540B~\cite{chung2022scaling} to annotate the quality of real model-generated summaries and utilized these annotated data as training data to tune a light-weight LLM (e.g., T5-11B) as a factual consistency summary evaluator, which predicts ``1" if the summary demonstrates factual consistency and ``0" otherwise. 
\citet{yue2023automatic} reused and repurposed the existing fact-checking, NLI, and summarization tasks datasets and obtained simulated data from open-domain QA datasets to tune light-weight LLMs for attribution evaluation, which generates attributable, extrapolatory or contradictory labels for the generated answer with given query and reference documents. 
\citet{li2023generative} created a dataset containing multiple scenarios and used GPT-4~\cite{openai2023gpt4} to generate evaluation judgments for each scenario as supervision signals to tune LLaMA as a generative evaluator, which can output overall quality rating for individual LLM-generated response in various scenarios. 
\citet{wang2023learning} repurposed existing datasets with proper anonymization and new personalized labels to tune LLaMA2~\cite{touvron2023llama} as a personalized story evaluation model which provides personalized evaluation for generated texts through outputting a grade in [1, 10] and detailed reviews.
\citet{kim2023prometheus} prompted GPT-4 to construct training data, including reference answers and crafted diverse customized score rubrics, and used them to tune LLaMA to evaluate model-generated responses of given instruction, which is generalized to realistic user demands.
\citet{ke2023critiquellm} instructed GPT-4 to collect referenced and reference-free training data with dialogue-based prompting, utilized to tune LLMs for evaluating the alignment of model-generated texts with human instructions through generating scores and explanations.
\cite{liu2023x} constructed a reference-free instruction-tuning dataset tailored for multi-aspect evaluation across summarization, dialogue and data-to-text tasks. Considering that there is an internal correlation between the evaluation aspects, this work tuned with auxiliary aspects additionally on diverse evaluation task forms. During inference, this work combined auxiliary and target aspects and predicted either the ``Yes'' or ``No'' label to judge whether the generated text satisfied the target aspect and compute the evaluation score.

\paragraph{Probability-based Evaluation.}
Some works train generative LLMs to calculate the generation probability of generated texts to evaluate text quality. For example,
\citet{thompson-post-2020-automatic} trained a transformer as a multilingual reference-to-candidate paraphraser to obtain the generated probability of model-generated translation based on their reference texts. 
\citet{qin2022t5score} tuned the T5 model in the generative and discriminative fashion, and used the probability of generating a text as the quality score.

\paragraph{Pairwise Evaluation.}
There are also some works tuning LLMs for comparison between generated text pairs.
\citet{wang2023pandalm} collected response pairs from LLMs and asked GPT-3.5 to generate output judgments, utilized which to tune LLaMA-7B to evaluate a pair of model-generated responses with the given query, accompanied by a concise description of the evaluation procedure. 
\citet{zheng2023judging} performed fine-tuning on Vicuna using a human votes dataset from Chatbot Arena to pairwise evaluate two answers with the given query.

\paragraph{Advanced Evaluation.}
Nearly all tuning-based evaluators are trained to emulate evaluate behavior (the score or explanations) produced by strong closed models like GPT-4 or ChatGPT. In the context of supervised fine-tuning, the majority of studies gravitate towards \textbf{\emph{holistic evaluation}}~\cite{li2023generative,wang2023pandalm,wang2023learning,kim2023prometheus}, which involves a comprehensive assessment of the generated content, providing an overarching explanation for the assigned score. It takes into account a diverse range of factors, including coherence, relevance, and fluency, to offer a holistic understanding of the quality of the hypothesis text. Besides, some studies explore \textbf{\emph{error-oriented evaluation}} which focused on examining and explaining the specific errors in the hypothesis text, offering insights into why a particular score is derived. This category delves into the fine-grained aspects of generated content to identify and justify evaluation outcomes.
For instance,
\citet{yue2023automatic} first defined different types of attribution errors, and then explored prompting LLMs or fine-tuning smaller LLMs on simulated and repurposed data from related tasks such as question answering (QA), fact-checking, natural language inference (NLI), and summarization. \citet{xu2023instructscore} utilized GPT-4 to construct fine-grained analysis data to tune LLaMA to generate error analysis for generated text compared with reference text, after which this work utilized real model-generated response-reference pairs to refine and self-train evaluator.
Furthermore, \citet{jiang2023tigerscore} sampled data from diverse text generation datasets, including summarization, translation and data2text, whose system outputs included real-world system output and GPT-4 synthesis, and prompted GPT-4 to curate error analysis to tune LLaMA for fine-grained evaluation.

\section{Benchmarks and Tasks}
\label{sec:benchmark}

LLM-based evaluators have found application across various NLG tasks. Simultaneously, a multitude of existing and recently introduced meta-evaluation benchmarks serve the purpose of validating the efficacy of these evaluators.
These benchmarks incorporate human annotations gauging the quality of generated text, and evaluating the degree of concurrence between automatic evaluators and human preferences.  Categorized based on the tasks involved, these benchmarks can be classified into single-scenario examples, such as machine translation and summarization, as well as multi-scenario benchmarks. This section will provide an overview of these NLG tasks and their associated meta-evaluation benchmarks.

\paragraph{Machine Translation (MT).} 
MT task is centered around converting a sentence or document from a source language into a target language while preserving the same semantic meaning. The Annual WMT Metrics Shared tasks~\cite{mathur-etal-2020-results,freitag-etal-2021-results,freitag-etal-2022-results} annually introduce a set of benchmarks encompassing model-generated translations, source text, reference text, and human judgment across multiple languages, such as English to German, English to Russian, among others. These benchmarks provide a valuable resource for evaluating the correlation between automatic evaluators and human judgment. Simultaneously, \citet{freitag-etal-2021-experts} curated and annotated outputs from 10 translated systems for both English-to-German and Chinese-to-English translation pairs in the WMT 2020 news translation task~\cite{barrault-etal-2020-findings}. Employing professionals and crowd workers as annotators, they assigned scalar ratings on a 7-point scale to each translation, utilizing multi-dimensional quality metrics scoring.

\paragraph{Text Summarizing (TS).} TS involves generating a concise and coherent summary of a given piece of text while capturing its essential meaning. 
There are many meta-evaluation benchmarks are proposed~\cite{grusky-etal-2018-newsroom,gliwa2019samsum,bhandari-etal-2020-evaluating,wang-etal-2020-asking,pagnoni-etal-2021-understanding,laban-etal-2022-summac,skopek2023towards}. 
One of the widely used benchmarks is SummEval~\cite{fabbri2021summeval}. This benchmark includes summaries generated by 16 models from 100 source news articles randomly sampled from the CNN/DailyMail test set~\cite{hermann2015teaching}, and each summary underwent annotation by five separate crowd-sourced workers and three independent experts on a Likert scale from 1 to 5 along four dimensions: coherence, consistency, fluency and relevance. 
In addition, \citet{shen2023opinsummeval} presented a meta-evaluation benchmark for opinion summarization tasks, including human judgments and outputs from 14 opinion summarization models over four dimensions: aspect relevance, self-coherence, sentiment consistency and readability, where opinion summarization task focuses on extracting opinions from a large number of reviews.

\paragraph{Dialogue Generation (DG).} DG task aims to generate human-like responses in the context of a conversation, including open-domain and task-oriented dialogue generation tasks. The model-generated dialogue should be natural and interesting, while also being consistent with the context.
\citet{mehri-eskenazi-2020-usr} performed human annotations across two open-domain dialog
corpora Topical-Chat~\cite{Gopalakrishnan2019TopicalChatTK} and PersonaChat~\cite{zhang2018personalizing}. For each dataset, 60 dialogue contexts are sampled with six responses per context for Topical-Chat and five responses for PersonaChat, where each response is generated from dialogue systems and human outputs. Each response is scored from 6 dimensions including naturalness, coherence, engagingness, groundedness, understandability and overall quality.
\citet{mehri-eskenazi-2020-unsupervised} sampled and annotated a subset from a set of conversations between a human and another human or two open-domain dialog systems~\cite{adiwardana2020towards}. Turn-level and dialog-level human judgment are performed, respectively, for each sampled conversation on eighteen dialog quality dimensions.

\paragraph{Image Caption (IC).} The task involves generating textual descriptions or captions for images. Meta-evaluation benchmarks of image caption contain human annotations for image-textual pairs~\cite{aditya2015images,vedantam2015cider}. 
For example, the commonly used Flickr 8k dataset \cite{hodosh2013framing} collects two sets of human annotations. One set includes 17K expert judgments annotation, which asks human experts to rate the image-caption pairs with scores ranging from 1 to 4, and another set includes 145K binary quality judgments gathered from CrowdFlower for each image-caption pair, which decide whether a caption describes the corresponding image or not.
Considering some NLG evaluators can only handle textual modal information, some meta-evaluation benchmarks also include a reference caption for each image.
\citet{cui2018learning} collected human judgments for twelve submission entries from the 2015 COCO Captioning Challenge on the COCO validation set~\cite{vinyals2016show}, where each system generates one caption for each image, and each image has five reference captions.

\paragraph{Data-to-Text (D2T).} D2T task involves generating fluent and factual human-readable text from structured data. 
\citet{mairesse-etal-2010-phrase} proposed BAGEL, which contains 202 samples about restaurants in Cambridge, where each sample includes structured information context with corresponding generated texts, references and human judgments. \citet{wen-etal-2015-semantically} further proposed SFRES and SFHOT, which contain 581 samples of restaurants and 398 samples of hotels in San Francisco, respectively. The human judgments in these benchmarks focus on informativeness, naturalness and overall quality of generated texts. 
WebNLG+ Shared Tasks~\cite{castro-ferreira-etal-2020-2020} also publish WebNLG dataset annually, which contains Wikipedia triples with corresponding human-annotated texts.

\paragraph{Story Generation (SG).} The task involves creating coherent and contextually relevant narratives or stories with the given beginning of a story or writing requirement. Most meta-evaluation benchmarks of story generation always contain stories and corresponding manually annotated judgment scores~\cite{guan-etal-2021-openmeva,chen-etal-2022-storyer}.
Besides, \citet{wang2023learning} created two personalized story evaluation benchmarks denoted as Per-MPST and Per-DOC to evaluate the quality of generated stories with a given evaluator persona.
This work repurposes existing datasets~\cite{kar-etal-2018-mpst,yang-etal-2023-doc} through anonymizing and summarizing. Both them view multiple reviews by the same reviewer as an implicit persona preference and provide personalized human judgements for each generated story.

\paragraph{General Generation (GE).}
As LLMs have been increasingly used in general NLG tasks, such as math, reason, dialogue and open-ended QA, etc., LLM evaluators have proposed to effectively evaluate the quality of the model-generated texts across multiple scenario~\cite{kim2023prometheus,ke2023critiquellm}. Accordingly, there are many multi-scenario meta-evaluation benchmarks are proposed~\cite{wang2023large,zheng2023judging,wang2023shepherd,yue2023automatic}. Typically, \citet{zhang2023wider} sampled 2,553 evaluation samples, including instructions and model-generated response pairs with corresponding human-annotated preference labels from multiple task datasets such as dialogue, open-domain QA, and programming.
Further, \citet{zeng2023evaluating} proposed a  benchmark that includes 419 evaluation samples and can be categorized into two parts: NATURAL and ADVERSARIAL sets. The former collects and filters instances from existing human-preference benchmarks to ensure that each instance has an objective preference. The latter includes the adversarial instances created by authors that go against instruction but have good superficial qualities and are challenging for evaluators.
\citet{liu2023alignbench} sampled 400 evaluation instances, including Chinese queries, corresponding references and model-generated answers from ALIGNBENCH across extensive task categories, such as open-ended questions, writing ability, logical reasoning, etc. Then the authors assigned human annotators to judge ratings for each instance to verify the agreement of LLM-based evaluators with human judging.

\section{Comparison with Traditional Evaluators}
\label{sec:comparison}

\begin{table*}[!t]
\centering
\resizebox{\linewidth}{!}{
\begin{tabular}{lcccccccccccccc}
\toprule
\multicolumn{1}{c}{\multirow{2}{*}{\textbf{Metrics}}} & \multicolumn{1}{c}{\multirow{2}{*}{\textbf{Sup}}} & \multicolumn{5}{c}{\textbf{SummEval}} & \multicolumn{5}{c}{\textbf{Topical-Chat}} & \multicolumn{3}{c}{\textbf{WMT22}} \\ 
\cmidrule(r){3-7}  \cmidrule(r){8-12} \cmidrule(r){13-15} 
                    &   & COH & CON & FLU & REL & Avg & NAT & COH & ENG & GRO & Avg & En-De & En-Ru & Zh-Eu
                       \\ \midrule
\multicolumn{15}{c}{{\textbf{Traditional Metrics (Word Overlap)}}} \\
\midrule
\texttt{ROUGE-1} & & 0.167 & 0.160 & 0.115 & 0.326 & 0.192 & 0.158 & 0.206 & 0.319 & 0.264 & 0.233 & - & - & - \\
\texttt{ROUGE-2} & & 0.184 & 0.187 & 0.159 & 0.290 & 0.205 & 0.168 & 0.247 & 0.337 & 0.311 & 0.266 & - & - & - \\
\texttt{ROUGE-L} & & 0.128 & 0.115 & 0.105 & 0.311 & 0.165 & 0.145 & 0.205 & 0.306 & 0.293 & 0.237 & - & - & - \\
\texttt{BLEU} & & - & - & - & - & - & 0.175 & 0.235 & 0.316 & 0.310 & 0.259 & 0.169 & 0.140 & 0.145 \\
\midrule
\multicolumn{15}{c}{{\textbf{BERT-based Metrics}}} \\
\midrule
\texttt{BERTScore} & & 0.284 & 0.110 & 0.193 & 0.312 & 0.225 & 0.209 & 0.233 & 0.335 & 0.317 & 0.273 & 0.232 & 0.192 & 0.316 \\
\texttt{BLEURT} & \checkmark & - & - & - & - & - & - & - & - & - & - & 0.344 & 0.359 & 0.361 \\
\texttt{BARTScore} & \checkmark & 0.448 & 0.382 & 0.356 & 0.356 & 0.385 & -0.053 & -0.079 & -0.084 & -0.197 & -0.103 & - & - & 0.220 \\
\texttt{UniTE} & \checkmark & - & - & - & - & - & - & - & - & - & - & 0.369 & 0.378 & 0.357 \\
\texttt{UniEval} & \checkmark & 0.575 & 0.446 & 0.449 & 0.426 & 0.474 & 0.450 & 0.616 & 0.615 & 0.590 & 0.568 & - & - & - \\
\midrule
\multicolumn{15}{c}{{\textbf{LLM-based Metrics}}} \\
\midrule
\texttt{GPTScore} & & 0.434 & 0.449 & 0.403 & 0.381 & 0.417 & - & - & - & - & - & - & - & 0.187 \\

\texttt{CHATGPT(DA)} & & 0.451 & 0.432 & 0.380 & 0.439 & 0.425 & 0.474 & 0.527 & 0.599 & 0.576 & 0.544 & 0.306 & 0.332 & 0.371 \\
\texttt{G-Eval} & & 0.582 & 0.507 & 0.455 & 0.547 & 0.514 & 0.607 & 0.590 & 0.605 & 0.536 & 0.590 & - & - & - \\
\texttt{Embed Llama} & & - & - & - & - & - & - & - & - & - & - & 0.400 & 0.227 & 0.217 \\
\texttt{X-Eval} & \checkmark & 0.530 & 0.428 & 0.461 & 0.500 & 0.480 & 0.478 & 0.622 & 0.593 & 0.728 & 0.605 & - & - & - \\

\bottomrule
\end{tabular}
}
\caption{Performance of traditional and LLM-based metrics on Text Summarizing (SummEval), Dialogue Generation (Topical-Chat) and Machine Translation (WMT22) tasks. We demonstrate the sample-level Spearman correlations on SummEval and Topical-Chat benchmarks and the segment-level Kendall-Tau correlations on WMT22 benchmarks respectively. \textbf{Sup} indicates the metric is supervised. The specific representation of the evaluation aspects (COH/CON/FLU/REL/NAT/ENG/GRO) is shown in Table ~\ref{tab:auto_metric_adoption}.
}
\label{tab:all}
\end{table*}

\paragraph{Qualitative Comparison}
Traditional evaluation metrics (e.g., BLEU~\cite{Papineni2002Bleu} and ROUGE) focus on exacting n-gram matches, which penalizes semantically correct but lexically different hypotheses. These methods are simple and fast but not robust to paraphrasing. BERTScore~\cite{Zhang2020BERTScore} measures quality through semantic similarity based on BERT contextual embeddings, effectively handling paraphrases and synonyms. However, such matching-based evaluators depend on the quality of the pre-trained embeddings, may struggle with very fine-grained semantic distinctions, and neglect the overall semantics of the hypotheses and references. Additionally, neither metric accounts for fluency or readability during evaluation and both still rely on reference texts.

In contrast, LLMs have a strong capability for language understanding and generation, which supports evaluating quality without needing references. They can adapt to various domains and languages, making them suitable for a wide range of NLG tasks without requiring task-specific feature engineering. LLMs also provide more nuanced evaluation criteria beyond traditional metrics, such as semantic coherence, fluency and possible explanations. However, LLM-based methods are computationally more intensive due to their vast architectures. Additionally, prompting LLMs for NLG evaluation requires careful crafting of prompts. Variations in these prompts can lead to substantial differences in evaluation outcomes, as indicated in \cite{gao2023human}. Section~\ref{sec:future} summarizes more open problems of LLM-based metrics.

\paragraph{Performance Comparison} Table \ref{tab:all} summarizes the performance of both traditional word-overlap metrics, BERT-based metrics and recent LLM-based metrics on representative benchmarks such as SummEval, WMT, and Topical-Chat. 
We can easy to observe that the latter two metrics generally perform better than word-overlap metrics.
Despite not being fine-tuned, the most competitive LLM-based methods (e.g., G-Eval for summarization and CHATGPT(DA) for machine translation) generally achieve a higher correlation with all traditional metrics, whether for unsupervised or fine-tuned methods. 
These results reveal the strong capability of LLMs in language understanding, contextual analysis, coherence checking, and fluency assessment of generated text.
Among the three tasks, the performance gap between LLM-based evaluators and traditional evaluators is not significant in the machine translation task. This phenomenon might be due to the limitations of LLM-based models in cross-lingual understanding. Additionally, according to the results of last row in the table, we can observe that the performance of different LLM-based metrics varies significantly, which implies their sensitivity to prompt crafting. In contrast, traditional unsupervised methods like ROUGE, BLEU, and BERTScore are more robust, although their overall performance is relatively worse.

\paragraph{Efficiency Comparison}
Table \ref{tab:infer_time} presents the average number of texts evaluated per second for different metrics in the SummEval (TS task) and Topical-chat (DG task) benchmarks. This comparison highlights the efficiency differences between traditional metrics and LLM-based metrics. Our tests were conducted on an NVIDIA A40 GPU. The results show that efficiency generally correlates with model size and traditional word-overlap metrics (e.g., BLEU and ROUGE) are significantly faster than other metrics. Specifically, LLM-based evaluators are about $200$ to $400$ times slower than traditional word-overlap metrics. However, their efficiency can be improved with advanced LM inference tools such as vLLM\footnote{{\url{https://github.com/vllm-project/vllm}}}. 
While LLM-based evaluators are suitable for offline evaluation, they may not be feasible for online evaluation.

\begin{table}[t!]
    \small
    \centering
    \begin{tabular}{lccc}
    \toprule
        Methods & Backbone &  TS & DG \\ \midrule
        BLEU  & - & \underline{977.31} & \underline{2344.16} \\
        ROUGE & - & 446.36 & 2379.24 \\
        BERTScore & BERT &  37.64 & 42.37 \\
        \midrule
        ChatGPT(DA) & ChatGPT & 1.94 & 1.87 \\
        G-Eval & GPT-4 & 1.51 & 1.40 \\
        TIGERScore & Llama & 2.67 & 3.72 \\
    \bottomrule 
    \end{tabular}
    \caption{Efficiency Comparison. We report the aver-
age number of texts evaluated per second for different metrics.}
    \label{tab:infer_time}
\end{table}




\section{Challenges and Open Problems}
\label{sec:future}
This paper provides a comprehensive review of recent natural language generation evaluations based on LLMs, encompassing both prompt-based and tuning-based evaluation approaches. Despite significant efforts and notable achievements across various text generation benchmarks, several challenges in the field persist.

\paragraph{Bias of LLM-based Evaluators.}
The use of LLMs as evaluators inherently cast the text evaluation as a generation task. Consequently, when LLMs are employed in this evaluator role, they may carry over biases intrinsic to their function as generators. These biases may include social biases, such as stereotypes related to specific demographic identities (e.g., race, gender, religion, culture, and ideology)~\cite{sheng2021societal}.
In addition to these general biases, LLMs-as-evaluators are subject to specific biases unique to their evaluative role. These include order bias, where preference is given to options based on their sequence~\cite{zheng2023judging, wang2023large, koo2023benchmarking}; egocentric bias, where a tendency exists to favor texts generated by the same LLM~\cite{liu2023llms, koo2023benchmarking}; and length bias, which leads to a preference for longer or shorter texts~\cite{zheng2023judging, koo2023benchmarking}.
Therefore, in leveraging LLMs for evaluation purposes, it is crucial to calibrate both the inherent biases of LLMs as well as those biases specific to their function as evaluators. This dual consideration is essential for the effective and fair use of LLMs in NLG evaluation tasks.

\paragraph{Robustness of LLM-based Evaluators.}
Most LLMs-based evaluation methods rely heavily on prompt engineering. However, the process of prompting LLMs
for NLG evaluation demands careful and meticulous crafting of prompts. The variations in these prompts can potentially lead to substantial differences in the outcomes of the evaluation process.
Some works have investigated the robustness of LLM-based evaluators by constructing adversarial datasets. These datasets are designed to test the evaluators' resilience by introducing false or off-topic information, thereby examining the impact of such distortions on their evaluative accuracy.
Their findings shed light on the significant room for improvement in the robustness of LLM-based evaluators. For instance, ~\citet{liu2023evaluate} developed two adversarial meta-evaluation datasets for dialogue generation with adversarial instances inconsistent with gold references. 
\citet{koo2023benchmarking} introduced a benchmark containing two adversarial aspects: Distraction and Bandwagon Effect, which involve appending irrelevant information or fabricated statistics, such as a misleading majority preference, to the initial instructions. The results suggest a general lack of robustness in many LLMs under such adversarial conditions. The robustness of LLM-based evaluators emerges as a critical area of exploration, underscoring the need for further research to enhance their robustness in the face of challenging or misleading inputs.

\paragraph{Which came first, the chicken or the egg?}
LLM-based evaluators frequently utilize GPT-4~\cite{liu2023gpteval, xu2023instructscore, zheng2023judging}, owing to its status as one of the most advanced LLM~\cite{openai2023gpt4}. However, relying on GPT-4 for evaluation might lead to biased or skewed results, especially when evaluating outputs generated by itself or an equally powerful model~\cite{bai2023benchmarking, zheng2023judging}. The impartiality of such evaluations is questionable if the evaluator (LLM-as-evaluator) possesses capabilities comparable to the model being evaluated (LLM-as-generator). This is compounded by the egocentric bias of LLMs, including GPT-4, to exhibit biases like favoring their own generated responses~\cite{bai2023benchmarking}. This scenario mirrors the chicken-and-egg dilemma: an LLM-based evaluator relies on a more powerful LLM, yet the development of a more powerful LLM depends on having a robust evaluator. To address this dilemma, a broader spectrum of evaluation methods is necessary, involving various benchmark~\cite{srivastava2022beyond, liang2022holistic}, evaluation criteria~\cite{sellam-etal-2020-bleurt}, and human feedback~\cite{xu2023instructscore, ouyang2022training} to ensure more balanced and comprehensive assessments.

\paragraph{Domain-Specific Evaluation.}
LLMs have been prevalent across various domains, such as  law~\cite{cui2023chatlaw}, medicine~\cite{singhal_large_2023}, finance~\cite{yang2023fingpt}, etc. However, most LLMs employed as evaluators are designed for general domains and are not specifically tailored to any particular field. This lack of specialization poses significant challenges. On one hand, these LLMs often lack the requisite domain-specific knowledge, making it difficult for them to accurately assess the correctness of content within specialized fields. On the other hand, the evaluation prompts need to be meticulously designed for different domains. This may involve tailoring the aspects of evaluation relevant to each field. For example, while evaluating legal documents, aspects such as legal accuracy and adherence to the judicial system are crucial~\cite{cui2023survey}, which differ significantly from the aspects relevant in medical or financial texts. Therefore, to enhance the efficacy of LLMs as evaluators in specialized domains, there's a pressing need to develop models that are not only domain-aware but also equipped with the capability to evaluate based on domain-specific criteria.


\paragraph{Unified Evaluation.}
LLMs have been expanded w.r.t their broad capabilities beyond traditional single-task focuses, encompassing complex instructions like coding and open-ended real-world requirements~\cite{openai2023gpt4, Significant_Gravitas_AutoGPT}. Consequently, there is an increasing demand for more comprehensive and flexible evaluation methods. 
However, traditional evaluation methods and most current LLM-based evaluators are limited to constrained tasks and evaluation aspects (cf. Table \ref{tab:auto_metric_adoption}). Some promising attempts have been made in this direction. For instance, MT-Bench~\cite{zheng2023judging} uses GPT-4 as an evaluator across multiple domains for multi-turn questions. Yet, this is too confined to a few evaluation aspects and limits dialogue to two turns only. Another model, Auto-J~\cite{auto-j}, approaches from a data construction perspective, training a 13B LLM on user queries and GPT-4 generated responses across a wide range of real-world scenarios. It accommodates diverse evaluation protocols and has been validated in 58 different scenarios, even outperforming many proprietary LLMs.
In light of increasingly complex user queries, we advocate that developing a more unified and contemporaneous evaluation protocol is a promising direction. Additionally, constructing high-quality, comprehensive datasets to train unified models holds great potential. Such advancements could significantly contribute to more effective and universal evaluations of LLMs.
\section{Conclusion}
\label{sec:conclusion}

In this paper, we have meticulously surveyed the role of LLMs in the evaluation of NLG. Our comprehensive taxonomy classifies works along three primary dimensions: evaluation function, evaluation references and evaluation task. This framework enabled us to systematically categorize and understand LLM-based evaluation methodologies. We delved into various LLM-based approaches, scrutinizing their strengths and comparing their differences. 
Additionally, we summarized prevalent meta-evaluation benchmarks for NLG evaluation. Throughout our study, we highlighted both the advancements and the prevailing challenges in this rapidly evolving field. While LLMs offer groundbreaking potential in evaluating NLG outputs, there still remain unresolved issues that require attention, including bias, robustness, the integration of hybrid evaluation methods, and the need for domain-specific and unified evaluation within LLM-based evaluators. We anticipate that addressing these challenges will pave the way for more general, effective, and reliable NLG evaluation techniques. Such advancements would contribute significantly to the progression of both NLG evaluation and the broader application of LLMs.


\bibliography{custom}
\bibliographystyle{acl_natbib}

\appendix



\end{document}